 \documentclass[final,5p,times]{elsarticle}
\usepackage{amssymb}
\makeatletter
\def\ps@pprintTitle{%
  \let\@oddhead\@empty
  \let\@evenhead\@empty
  \let\@oddfoot\@empty
  \let\@evenfoot\@oddfoot
}
\makeatother
\usepackage{multirow}
\usepackage{amsmath}
\usepackage{graphicx}
\usepackage{epstopdf} %converting to PDF
\usepackage{float}
\usepackage{color}
\usepackage[dvipsnames]{xcolor}
\usepackage{balance}
\usepackage{soul}
\usepackage{mathrsfs} 
\usepackage{array}
\usepackage{rotating}
\usepackage{pbox}
\usepackage{subfig}
\newcolumntype{C}[1]{>{\let\newline\\\raggedright\hspace{0pt}}m{#1}}
\usepackage{algorithm}% http://ctan.org/pkg/algorithmicx
\usepackage[noend]{algpseudocode}
\usepackage{threeparttable} 
\usepackage{multicol}
\newcommand\blfootnote[1]{%
  \begingroup
  \renewcommand\thefootnote{}\footnote{#1}%
  \addtocounter{footnote}{-1}%
  \endgroup
}
\begin{document}

\begin{frontmatter}



\title{MAP-Elites based Hyper-Heuristic for the Resource Constrained Project Scheduling Problem }


\author{Shelvin Chand$^{a,*}$, Kousik Rajesh$^{b,*}$, Rohitash Chandra$^{c}$}

\address[1]{Commonwealth Scientific and Industrial Research Organisation, Brisbane, Australia}
\address[2]{Indian Institute of Technology, Guwahati, India}
\address[3]{University of New South Wales, Sydney, Australia}

\begin{abstract}
The resource constrained project scheduling problem (RCPSP) is an NP-Hard combinatorial optimization problem. The objective of RCPSP is to schedule a set of activities without violating any activity precedence or resource constraints. In recent years researchers have moved away from complex solution methodologies, such as meta heuristics and exact mathematical approaches, towards more simple intuitive solutions like priority rules. This often involves using a genetic programming based hyper-heuristic (GPHH) to discover new priority rules which can be applied to new unseen cases.  A common problem affecting GPHH is diversity in evolution which often leads to poor quality output. In this paper, we present a MAP-Elites based hyper-heuristic (MEHH) for the automated discovery of efficient priority rules for RCPSP. MAP-Elites uses a quality diversity based approach which explicitly maintains an archive of diverse solutions characterised along multiple feature dimensions. In order to demonstrate the benefits of our proposed hyper-heuristic, we compare the overall performance against a traditional GPHH and priority rules proposed by human experts. Our results indicate strong improvements in both diversity and performance. In particular we see major improvements for larger instances which have been under-studied in the existing literature. 
\end{abstract}


\end{frontmatter}

\section{Introduction}
\blfootnote{* These authors contributed equally to the paper}
% \blfootnote{E-mail addresses: \href{mailto:shelvin.chand@csiro.au}{shelvin.chand@csiro.au (S. Chand)}, \href{mailto:kousik18@iitg.ac.in}{kousik18@iitg.ac.in (K. Rajesh)}, \href{mailto:rohitash.chandra@unsw.edu.au}{rohitash.chandra@unsw.edu.au (R. Chandra)}}
The resource constrained project scheduling problem (RCPSP) is an NP-Hard combinatorial optimization problem where the goal is to schedule a set of activities without violating any of the activity precedence or resource constraints \cite{kolisch1996a}. Computational tools for solving RCPSP are well sought after due to its wide applicability in areas such as manufacturing \cite{kolisch1996, li2018efficient}, construction and software development \cite{brucker1999resource}.

Recently there has been a surge in the use of genetic programming based hyper heuristics (GPHH) \cite{CHAND2018146, DUMIC2018211} for discovering efficient priority rules for RCPSP. Compared to traditional metaheuristics \cite{Zamani2013, Debels2007} and exact methods, these priority rules \cite{pritsker1969}, provide a much more simple, intuitive and computationally inexpensive scheduling approach that is well suited for dynamic environments. GPHH takes a set of problem characteristics/features as input and uses the process of evolution and recombination to construct useful mathematical expressions representing priority rules \cite{CHAND2018146, Nguyen2013}.  These rules can then be used to decide the order in which activities are processed or scheduled. 
Though metaheuristics \cite{PELLERIN2020395, proonjin, goncharovlenov, 10.1007/11424925_41} may generate better-performing schedules, the computational costs often make them infeasible for quick generation of schedules. The best performing metaheuristic \cite{proonjin} uses genetic algorithms with neighborhood search. Each problem instance requires more than 1000 iterations to arrive at a good solution. In contrast, hyper-heuristics provide a one-shot solution. Once the operator tree of GPHH is evolved, it can be used to quickly generate a well-performing schedule for any instance size. Thus, hyper-heuristics provide a more scalable and generalizable approach to RCPSP.

Most GPHHs within literature follow a standard approach where they are initialized with a set of randomly generated solutions and guided using an elitist approach. The best performing solutions are carried forward across generations and used as 'parents' for generating new, hopefully better, solutions. However, these hyper-heuristics often lose diversity within the population quite early and suffer from pre-mature convergence and over-fitting problems \cite{CHAND2018146, Branke2016}. Essentially, the elitist approach may often result in insufficient exploration of the search space. When this is combined with the bloat effect \cite{affenzeller2014gaining}, a common problem affecting GPs, it often leads to poor quality output in terms of the evolved rules.  

One way to fix this problem is by using a quality-diversity approach such as the Multidimensional Archive of Phenotypic Elites (MAP-Elites) algorithm \cite{mouret2015illuminating}. In comparison to traditional elitist methods, MAP-Elites maintains a map or grid of high performing solutions which are characterised along multiple features of interest as defined by the decision maker \cite{mouret2015illuminating}. While traditional elitist methods output a single best solution, MAP-Elites outputs a set of diverse, high-performing solutions \cite{mouret2015illuminating}. This allows the decision maker to not only have a wider range of choice but also to gain a better understanding of the search space. But perhaps most importantly, in defining a feature map, the decision maker is able to guide the search using domain specific knowledge. In fact, prior research has shown that since MAP-Elites explores a larger part of search space, it also tends to find better solutions than traditional elitist methods \cite{mouret2015illuminating}. 

%maybe a paragraph complaining about lack of research in bigger instances of the problem which further motivates our wok

In this paper, we study the automated evolution of priority rules for RCPSP using a MAP-Elites based hyper-heuristic. In order to demonstrate the benefits of our proposed hyper-heuristic, we compare the overall performance against a traditional GPHH. We also compare the discovered rules against the state of the art rules proposed by human experts. To the best of our knowledge, this is first study looking into the use of a MAP-Elites based hyper-heuristic in evolving rules for RCPSP. We also focus on larger instance sizes which have been rarely studied in existing literature. More specifically, the key intended contributions of this paper are:
\begin{itemize}
    \item Development and study of a quality-diversity hyper-heuristic framework to evolve priority rules for the classical RCPSP.
    \item In-depth analysis on quality-diversity elements such as MAP-Elites feature set, grid size, etc.
    \item Performance characterization of the evolved rules and benchmarking with the existing rules in the literature on a large set of instances with as many as 300 activities. 
\end{itemize}

The remainder of the paper is organized as follows. Section 2 provides a discussion on existing literature. Section 3 gives details on the proposed hyper-heuristic approach for evolving rules. Section 4 provides a discussion on the results obtained while Section 5 summarizes the findings of this paper and highlights some future research directions.

\section{Background}
\subsection{Problem Definition}

The RCPSP~\cite{Kolisch1999} involves a project with $M$ activities which need to be scheduled while considering two types of constraints: 

\begin{itemize}
	\item \emph{Precedence constraints}:  These represent the interdependence between the activities. If activity $j$ is a successor of activity $i$ then activity $i$ must be completed before activity $j$ can be started. 
	\item \emph{Resource constraints}: These represent limitations on resources such as manpower, budget, etc. Each project is assigned a set of $K$ renewable resources where each resource $k$ is available in $R_{k}$ units for the entire duration of the project. Each activity may require one or more of these resources to be completed. While scheduling the activities, the daily resource usage for resource $k$ can not exceed $R_{k}$ units. 
\end{itemize}

Each activity $j$ takes $d_{j}$ time units to complete. The overall goal of the problem is to minimize the \emph{makespan}, i.e., the total duration from the start of first to the end of last scheduled activity.

\begin{figure}[H]
    \centering
    \includegraphics[width=9cm]{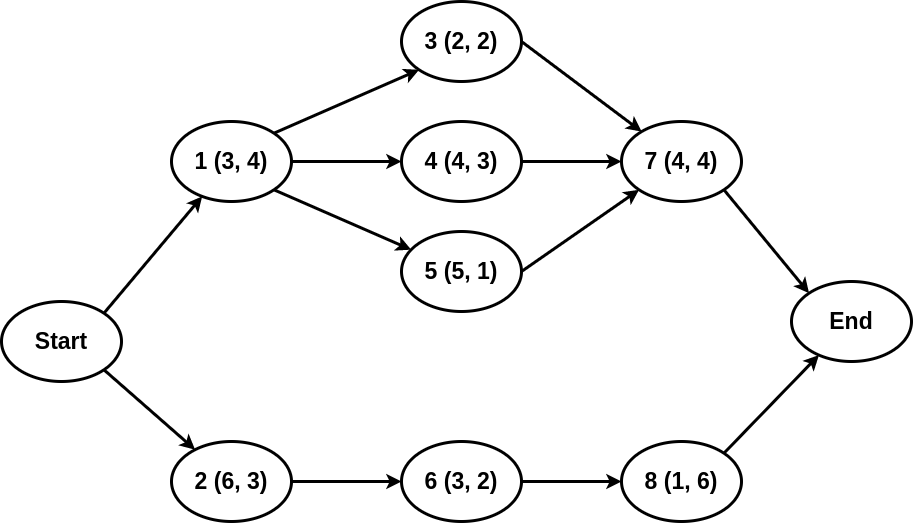}
    \caption{An RCPSP instance visualised as a Directed Acyclic Graph (DAG) with each node representing an activity. There is a single resource available (K = 1) and a node is represented as $i(d_i, r_i)$ where $d_i$ and $r_i$ represent the duration and resource requirement for activity $i$. The arrows represent the precedence constraints between activities.}
    \label{fig:aon_graph}
\end{figure}

\begin{figure}[htbp!]
    \centering
    \includegraphics[width=9cm]{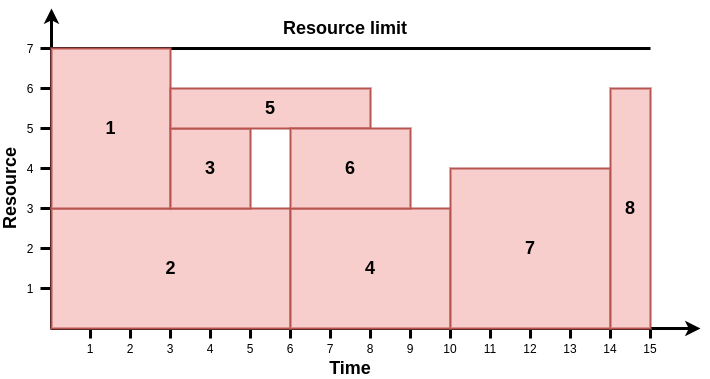}
    \caption{A possible schedule for the RCPSP instance shown in Fig \ref{fig:aon_graph}.}
    \label{fig:aon_schedule}
\end{figure}

An example activity on node (AON) diagram for an RCPSP instance is given in Fig \ref{fig:aon_graph}. A possible schedule for this project that doesn't violate the precedence and resource constraints is shown in Fig \ref{fig:aon_schedule}.

The RCPSP is considered to be NP-hard \cite{BLAZEWICZ198311} and therefore the exact methods \cite{Carruthers1966} are only practical for smaller instances. For larger instances, heuristic/meta-heuristic methods such as genetic algorithm \cite{ZAMANI2013552}, particle swarm optimization \cite{KOULINAS2014680}, simulated annealing \cite{VALLS2005375} and hybrid search \cite{ASTA2016476}, etc. are utilized for finding 'good-enough' or 'near-optimal' solutions. 

\subsection{Quality-Diversity based Hyper-Heuristics}
The literature on quality-diversity based hyper-heuristics is rather scant. There are a number of studies such as \cite{wanggecco19, sunguy18, CHAND2019897}  which consider diversity in one way or another to improve evolution. However, there are very few studies that actually utilize a quality-diversity like framework. Nguyen et al. \cite{gng2018Su} proposed the adaptive charting genetic programming (ACGP) hyper-heuristic (for the dynamic flexible job shop scheduling problem) which incrementally maps the explored areas of the search space and adaptively distributes its search effort to efficiently explore the search space. Furthermore, growing neural gas (GNG) and principle component analysis (PCA) utilize the phenotypic characteristics of the evolved heuristics to efficiently generate and update the map. Finally, a surrogate assisted model utilizes this map to determine which heuristics are to be explored in the next generation. 

While ACGP presented promising results, it also has its fair share of challenges. Firstly, GNG and PCA introduce a number of extra parameters which can be difficult to tune. Furthermore, the mapping strategy seems less intuitive from the point of view of a decision maker since it is less structured as opposed to the way MAP-Elites constructs its archive. This may also lead to certain areas of the search space being left under-explored or totally ignored. A variant of this algorithm has also been applied to the resource constrained job scheduling problem \cite{gngrcjp}. The same authors \cite{pcenguyen2021} later presented an extension of this algorithm with improved diversity control, bloat control and human-driven interactive evolution.

\subsection{GPHH for RCPSP}
Priority rules are a much more preferred option, over meta-heuristics and exact methods, when it comes to scheduling in dynamic environments. Meta-heuristics and exact methods are usually time consuming to run and not to mention, that tuning the parameters for such methods is often non-trivial. But perhaps most importantly, these methods lack flexibility and generalization capability needed for real world problems \cite{CHAND2018146}. Priority rules are simple, intuitive and easy to use. They also allow for quick decision making which is especially important in dynamic environments where project elements are constantly changing and evolving.  Priority rules are also most often void of any complex parameters that need tuning. Given their obvious advantages, the use of priority rules for scheduling  complex projects is on the rise. However, designing such rules is often non-trivial. The literature contains a number of rules which have been proposed over the years, by domain experts. Some of these are outlined are in Table \ref{existingRules}. A more comprehensive list of rules can be found in \cite{KLEIN2000619}.

Human experts are limited in their ability and scope when it comes to proposing new and improved priority rules. As a result there has been a recent push towards automating this process. Genetic programming based hyper-heuristics (GPHH) are the most preferred option as they have had success in evolving mathematical expressions \cite{Huynh2018}.

To the best of our knowledge, the first study on evolving priority rules for RCPSP was by Frankola et al. \cite{Frankola2008}. The authors proposed a GPHH which evolved arithmetic priority rules constructed using binary operators and a limited set of activity attributes which mostly focused on resource usage and network structure. This would be seen more so an exploratory study since the authors did not present any in-depth analysis of the results or any detailed comparison with existing human designed priority rules. This was followed by a more in-depth study by \v{S}i{\v{s}}ejkovi{\'c} in \cite{Dom2016} in which he explored the use of a much wider attribute and operator set. The author first used a greedy feature selection approach to identify the best performing attributes. Later, these attributes were combined to form arithmetic priority rules. The resulting rules exhibited competitive performance but were unable to outperform the best priority rules from literature.

Chand et al. \cite{CHAND2018146} proposed a GPHH framework in which they explored the use of different rule representations and a more structured selection of attributes. The authors compared the classical arithmetic representation with a decision tree like representation similar to the one proposed in \cite{Nguyen2013}. They also proposed an attribute set that focused on three key aspects of any project, namely the resource requirements, the critical path and the precedence network. The resulting rules outperformed all other existing human designed rules including LST, WCS, ACS, etc. In terms of the study on representation types, the authors found that the arithmetic representation performed the best while the decision tree representation gave competitive results with greater understandability. 

The same authors \cite{CHAND2019897} later proposed a multi-objective GPHH which tried to optimize performance as well as rule complexity. All of the studies prior to this had only focused on static instances. This study for the first time focused on dynamic RCPSP in which resource availability varied with time. The results indicated strong improvements over both, existing human designed rules and also the results presented in \cite{CHAND2018146}.

To the best of our knowledge there have been no studies so far that deviate from utilizing a traditional elitist approach to a more quality-diversity based approach. Some of the studies mentioned above \cite{CHAND2018146, CHAND2019897} report cases of pre-mature convergence and over-fitting which can be attributed to some extent to the elitist nature of traditional GPs. Also, nearly all these studies only focus on smaller instances (30 to 120 activities). However, the true benefit of using a quality-diversity based approach over an elitist approach becomes much more apparent when we work on large instances. In this study we aim to explore the benefits of a quality-diversity based hyper heuristic which utilizes MAP-Elites to search for high performing priority rules across a structured feature map. We also specifically focus on evolving rules for larger instances (300 activities) which are much more closer to the type of instances encountered in real world project scheduling.
\color{black}
\begin{figure}[htbp!]
    \centering
    \includegraphics[width=9cm]{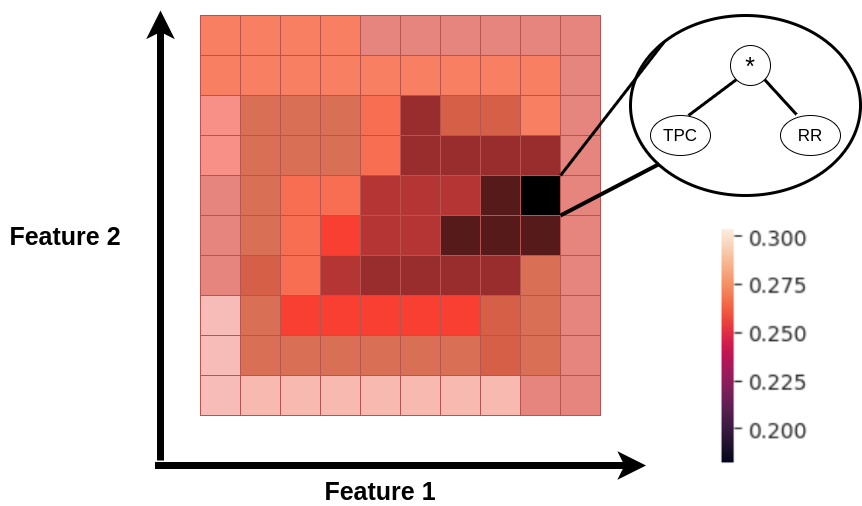}
    \caption{Solutions within a MAP-Elites grid.}
    \label{fig:grid_cs}
\end{figure}

\begin{table}[htbp!]\footnotesize
	
	\centering
	\caption{Some of the prominent existing priority rules.}
	
	\label{existingRules}
	\begin{threeparttable}
		\begin{tabular}{c c c c }
			\hline
			Rule                                & Abbrv. & Extr. & Description                      \\ \hline
			Early Start Time  \cite{Davis1975}                  & EST    & Min   & $ES_{j}$                         \\ 
			Early Finish Time  \cite{Davis1975}                 & EFT    & Min   & $EF_{j}$                         \\ 
			Late Start Time     \cite{Davis1975}                & LST    & Min   & $LS_{j}$                         \\ 
			Late Finish Time \cite{Davis1975}                   & LFT    & Min   & $LF_{j}$                         \\ 
			Most Total Successors  \cite{kolisch1996}             & MTS    & Max   & $|\bar{S_{j}}|$                   \\  
			First In First Out \cite{Davis1975}                   & FIFO   & Min  & $ ID_{j}$ 
			 \\
			 Shortest Processing Time \cite{Davis1975}            & SPT    & Min  & $d_{j}$
			 \\
			 
			Greatest Rank Position Weight\cite{kolisch1996}       & GRPW   & Max   & $d_{j}+ \sum_{i\in S_{j}}{d_{i}}$ \\ 
             Greatest Resource Demand \cite{Davis1975}           & GRD     & Max   & $ d_{j} * \sum_{k=1} ^{K} {r_{kj}}$ \\
             \hline
		\end{tabular}
		\begin{tablenotes}
			\item[1]  $j \in \{1...M\}$
			\item[2]$\bar{S_{j}}$ is the complete set of successor activities for activity $j$.
		\end{tablenotes}
	\end{threeparttable}
\end{table}

\begin{table*}[htbp!]\footnotesize
	\centering
	\setlength{\tabcolsep}{1pt}
	\caption{Operator Set}
	\label{operators}
	\begin{tabular}{|l|l|l|l|l|l|l|} 
		\hline
		Add(a,b)          & Mul(a,b)          & Sub(a,b)          & Div(a,b)          & Max(a,b)          &  Min(a,b)      &  Neg1(a)        \\ 
		\hline
		$a+b$      & $a*b$  &  $a-b$ &	 $\begin{cases} a/b \quad if\ b>0\\    0 \quad otherwise  \end{cases}$  & $\begin{cases} a \quad if\ a>b\\ b \quad otherwise	\end{cases}$  &  $\begin{cases} a \quad if\ a<b\\b \quad otherwise\end{cases}$  & $-1*a$  \\ \hline
	\end{tabular}
\end{table*}

\section{MAP-Elites for RCPSP}
Our proposed framework is composed of a number of components which include the MAP-Elites algorithm for evolving the rules, a set of operators (Table \ref{operators}) and attributes (Table \ref{activityAttributes}) for constructing the rules and a schedule generation scheme for generating complete schedules from the evolved rules. These are individually outlined below.

\subsection{MAP-Elites Hyper-Heuristic (MEHH)}
We build our hyper-heuristic using the original version of MAP-Elites (Algorithm \ref{map_elites_algorithm}) proposed by Mouret and Clune \cite{mouret2015illuminating}. The features used in constructing the feature map/grid are detailed in the next sub-section. The algorithm starts by generating a set of random trees. These trees are evaluated on the training set to determine their fitness. Next, their feature values are calculated and these solutions are placed into the feature grid. If two or more solutions fall within the same feature discretisation, then only the solution with the lowest fitness value is retained. 

Subsequently, crossover and mutation operators are applied to solutions within the grid in order to generate new solutions. These solutions are similarly evaluated and the grid is updated. This process is repeated until the termination condition is reached. All solutions within the final grid are evaluated on the validation set and the best performing solution is selected to be evaluated on the test set. 

The fitness of an individual solution is the average percentage deviation of the calculated makespan ($ms_i$) from the instance lower bound ($lb_i$). The lower bound used here is calculated using the critical path method, which generates a precedence feasible schedule while ignoring the resource constraints. The crossover and mutation operators are the same as the ones used in our previous work \cite{CHAND2018146, CHAND2019100556, CHAND2019897}.

\begin{algorithm}[H]
	%\scriptsize
	\caption{Basic MAP-Elites Algorithm}
	\label{map_elites_algorithm} %\scriptsize
	\begin{algorithmic}[1]
	 \For{$gen=0$ to $g$}
	        \For{$i=1$ to $\lambda$}
	            \If{$gen=0$}
	            \State $y \gets random\_individual()$
	            
	            \Else
	            \State $y \gets random\_variation(P)$
	            \EndIf
	            \State $f \gets calculate\_features(y)$
	            \State $p \gets calculate\_performance(y)$
	            \If{$P(f)=\Phi$ or $F(P(f)) > p$}
	            \State $P(f) \gets y$
	            \EndIf
	        \EndFor
	    \EndFor
	
	\end{algorithmic}
	\label{alg:mapelites}
\end{algorithm}

\subsection{MEHH Features}
A given combination of feature values correspond to one unique individual in the grid if the cell pointed to by the feature values is non empty. This is illustrated in Figure \ref{fig:grid_cs}. MAP-Elites was originally designed to operate on phenotypic features, as the name suggests, however in this paper we have used a combination of genotypic and phenotypic features. Our feature map/grid is constructed using the following features:
\begin{enumerate}
    \item \textbf{Number of nodes in an individual}: This is the total number of nodes in the tree representing the solution which include both internal (operations) and leaf (activity attribute) nodes.
    
    \item \textbf{Number of Resource related nodes}: These are the leaf nodes which carry information about resource usage of an activity; i.e $RR_j, AvgRReq_j, MaxRReq_j, MinRReq_j$.
    
    \item \textbf{Slack}: The slack provides a measure of how loosely packed the schedule is. The slack value for a given activity within a generated schedule is taken to be the maximum amount of time the activity can be delayed without causing a violation of resource or precedence constraints. The total for a given schedule is the sum of the individual activity slacks.
\end{enumerate}

Note that the first two features are properties of the generated tree itself, and independent of the performance on the instances in the training set. The third feature is dependent on the training set and the resulting schedules generated by the tree. In our experiments we limit the tree height. The number of nodes in a tree and the number of resource related nodes are bounded due to the fixed maximum height of generated trees; hence, these features are not normalised. The slack value depends on the number of instances and the size of the instances; hence, there is a need to normalise it.  The total slack of an individual $\zeta$ on the training set T is given by: 
\[
\text{Total slack} = \frac{1}{|T|}\sum_{i \in T} \frac{slack_i}{n_i}
\]
where $T$ denotes the training set, $slack_i$ is the slack value for the schedule generated on instance $i$ by the individual $\zeta$ and $n_i$ is the number of activities in instance $i$.

The resulting slack was found experimentally which fell in the range [1.65, 2.00]; hence, the feature domain is chosen appropriately. Slack calculation is outlined in Algorithm \ref{slack} and the variables involved are explained below:

\begin{itemize}
    \item \textit{slack} stores the sum of slack over all activities
    \item \textit{n} is the total number of activities in the instance.
    \item \textit{leastTime}  stores the least start time for the $i^{th}$ job over all its successors. The $i^{th}$ job has to end before the leastTime.
    \item \textit{startTimes} is an array where the $i^{th}$ element represents the start time of the $i^{th}$ job.
    \item \textit{finishTimes} is an array where the $i^{th}$ element represents the finish time of the $i^{th}$ job.
    \item \textit{jobRes} is an array where the $i^{th}$ element represents the resource required to complete the $i^{th}$ job.
    \item \textit{totalRes} is the amount of total resource available to the instance.
    \item \textit{resCons} represent an array for the amount of resource being consumed at time $j$.
    
\end{itemize}

\begin{algorithm}[htbp]
	%\scriptsize
	\small
	\caption{Calculating slack for a schedule }
	\label{slack} %\scriptsize
	\begin{algorithmic}[1]
		\State Set: $slack=0$
		\For{$i=1$ to $n-1$}
		\State Set : leastTime= $\infty$
		\For{$j$ in $successors[i]$} 
		\State $leastTime=min(leastTime,startTimes[j])$
        \EndFor
        \For{$j=finishTimes[i]+1$ to $leastTime-1$ }
            \State schedulable $\gets$ True
            \For{$k=1$ to $K$ }
            \If{$\textit{jobRes}[i][k] + \textit{resCons}[j][k] >  \textit{totalRes}[k] $}
                \State schedulable $\gets$ False
                \State break
            \EndIf
        \EndFor
        \If{schedulable}
        \State $slack \gets slack + 1$
         \Else
        \State break
       
        \EndIf
        \EndFor
        \EndFor
	\end{algorithmic}
\end{algorithm}

\color{black}

\subsection{Priority Rule Representation}
In this study, we utilize the widely used arithmetic representation \cite{Nguyen2013, CHAND2018146} for  candidate solutions within our hyper-heuristic. The arithmetic representation forms priority rules by combining arithmetic operators with activity attributes. The main advantage of arithmetic representation is the flexibility it provides in an evolutionary environment, allowing for easy recombination without resulting in infeasible tree structures \cite{CHAND2019100556}, as is the case with other forms of representations (eg. decision tree representation \cite{CHAND2018146}). We utilize six binary operators and one unary operator which are listed in Table \ref{operators}.

\subsection{Activity Attributes}

As mentioned previously, a candidate solution within our hyper-heuristic is a mathematical expression representing some combination of mathematical operators and activity attributes. We consider 10 different activity attributes in this research. The activity attributes that we consider are listed below, and their calculations are detailed in Table \ref{activityAttributes}:
\begin{itemize}
	\item \textit{TSC/TPC}: The total count of the (immediate and non-immediate) successors/predecessors of an activity.
	\item \textit{ES/EF}: The earliest start/finish time for an activity in the precedence feasible schedule calculated by relaxing the resource constraints where each activity is scheduled as early as possible. 
	\item \textit{LS/LF}: The latest start/finish time for an activity in the precedence feasible schedule calculated by relaxing the resource constraints where each activity is scheduled as late as possible using backward recursion~\cite{Elmaghraby1977}. 
	\item \textit{RR}: The total number of resources required by an activity.
	\item \textit{AvgRReq}: The average resource requirement of an activity.
	\item \textit{MinRReq}: The minimum resource requirement of an activity. 
	\item \textit{MaxRReq}: The maximum resource requirement of an activity. 
	
\end{itemize}

\begin{table}[t!]\footnotesize
	\centering
	\caption{Activity Attribute Set}
	\label{activityAttributes}
	\begin{threeparttable}
		\begin{tabular}{l|l}
			\hline
			\textbf{Attribute}           & \textbf{Details} \\ \hline
			Early Start~($ES_{j}$)             &    $\frac{1}{max(ES)} ES_{j}$             \\ 
			Early Finish~($EF_{j}$)            &    $\frac{1}{max(EF)} EF_{j}$           \\ 
			Late Start~($LS_{j}$)              &    $\frac{1}{max(LS)} LS_{j}$         \\ 
			Late Finish~($LF_{j}$)           &    $\frac{1}{max(LF)} LF_{j}$                 \\ 
			Total Predecessor Count~($TPC_{j}$)     &   $\frac{1}{M-1}|\bar{P_{j}}|$            \\ 
			Total Successor Count~($TSC_{j}$)       &   $\frac{1}{M-1}|\bar{S_{j}}|$              \\ 
			Resources Required~($RR_{j}$)         &    $\frac{1}{K}\sum_{k=1}^{K}\begin{cases} 
			1 \quad if \ r_{jk}>0\\    
			0 \quad otherwise   
			\end{cases}$              \\ 
			Avg. Resource Requirement~($AvgRReq_{j}$) & $\frac{1}{K}\sum_{k=1}^{K}\{\frac{1}{R_{k}} r_{jk} \}$                 \\ 
			Max Resource Requirement~($MaxRReq_{j}$) &  $Min(\frac{1}{R_{k}} r_{jk},\ k\in 1...K)$                \\ 
			Min Resource Requirement~($MinRReq_{j}$) &  $Max(\frac{1}{R_{k}} r_{jk},\ k\in 1...K)$                \\ \hline
		\end{tabular}
		\begin{tablenotes}
			\item[1] $\bar{P_{j}}$ is the complete set of predecessor activities for activity $j$.
		\end{tablenotes}
	\end{threeparttable}
\end{table}
These attributes have been used in our previous works \cite{CHAND2018146, CHAND2019100556} and have shown to be useful in constructing good priority rules. All attributes are also scaled between 0 and 1 to avoid any undesirable outcomes resulting from attributes with particularly high or low values.

\subsection{Schedule Generation Scheme}

There are two different schedule generation schemes detailed in \cite{KOLISCH1996320} which are widely used within the RCPSP literature. These are the parallel schedule generation scheme (PSGS) and the serial schedule generation scheme (SSGS) \cite{Kolisch1999}. It has been shown in previous research \cite{kolisch1996,CHAND2018146} that PSGS performs better when scheduling with priority rules. PSGS is  also based on time increment which makes it suitable for dynamic online scheduling \cite{CHAND2018146}.  Hence, for these reasons, we use PSGS within our hyper-heuristic framework. We use the implementation detailed in \cite{Kolisch1999}.

When generating a schedule with a priority rule, we always pick the activity with minimum value of the priority function to be scheduled next. If two activities have the same priority value, the tie is broken using their activity ID, that is, the one with the lower ID value is picked first.

\section{Experiment and Results}

\subsection{Experiment setup}
Our key algorithmic comparison is between a standard GPHH as detailed in \cite{CHAND2018146} and our proposed MEHH with varying archive sizes. Table \ref{table:gp_parameters} gives a summary of the algorithm parameters. These are standard values which have been used in our previous research \cite{CHAND2018146}. The feature domains have been determined using preliminary runs. Both methods were implemented using the DEAP \footnote{\url{https://deap.readthedocs.io/en/master/api/algo.html}} framework. All data and code from our research is available on github \footnote{\url{https://github.com/sydney-machine-learning/MEHH_RCPSP}}.

For all our experiments we use a training, validation and test set. For the training set we use the entire J30 instance set from the Project Scheduling Library (PSPLib) \cite{KOLISCH1997205}. J30 set consists of 480 instances with 30 activities each. The instances are constructed using three complexity parameters, namely network complexity (NC), resource factor (RF) and resources strength (RS). The instances were constructed using 48 different parameter combinations and each combination consists of 10 instances. For the validation and test, we use the RG300 instances.  The RG300 \cite{Debels2007} set was constructed using three complexity parameters, namely order strength (OS), resource utilization (RU), resource constrainedness (RC). The instances were constructed using 48 different parameter combinations and each combination consists of 10 instances. For validation, we pick the first instance from each parameter combination, resulting in 48(10\%) instances in total. The remaining 90\% of the instances are used for testing. Having an equal number of instances from each parameter combination ensures that instances of different hardness levels are included in training and validation. Using a validation set helps us reduce the effect of over-fitting and also lets us take advantage of the diversity in the archive. Training, validation and test performance is based on average percentage deviation from the lower bounds, with a lower value considered to be better.%\textcolor{OliveGreen}{The rationale behind choosing a training set composed of J30 instances and a test set consisting of RG300 instances is to test the generalizability of the solution, both these sets are composed of instances from very different distributions with respect to instance size, generation method and parameters. The RG300 instances were chosen for the test set to demonstrate the performance improvement of MEHH over GPHH on large instance sizes.}
%\textcolor{OliveGreen}{Evolved heuristics will be evaluated based on their performance on the test set, a lower percentage deviation or makespan is better and hence these will act as the response variables.}
% We also pick the Best, Worst and Median rules from these 31 to be further evaluated and analyzed. Each of the MAP-Elites rules is denoted as MpE$t_{<grid\ size >}$- {$<metric>$} where \textit{metric} can be either B (Best), W(worst) or M(median) while the $grid_size$ can be either 125, 1000, 3375, 8000. 

\begin{center}
\begin{table}[]
\centering
\small
\caption{Parameters for GPHH and MEHH.}
\label{table:gp_parameters}
\begin{tabular}{ l c } 
 \hline
 Parameter & Value \\ 
 \hline
 Population size & 1024 \\
 Stopping condition & 25 Generations \\
 Fitness function & Average \% Deviation\\
 Selection method & Tournament Selection \\
 Selection pool size & 7 \\
 Gen height & [2,5] \\
 Height limit & 7 \\
 Crossover probability & 0.8 \\
 Mutation probability & 0.2 \\
 Number of runs & 31 \\
 Max items per bin (MEHH) & 1 \\
 Fitness domain (MEHH) & [0,1]\\
 $Feature_1$ (\# of nodes) domain (MEHH) & [4,127]\\
 $Feature_2$ (\# of res. nodes) domain (MEHH) & [0, 30]\\
 $Feature_3$ (Slack) domain (MEHH) & [1.65, 2.00]\\
 
 \hline
\end{tabular}
\end{table}
\end{center}

\subsection{Comparison Against GP}
We start by comparing our proposed MEHH with a standard GPHH \cite{CHAND2018146}. For MEHH we consider 4 different variants in which everything is kept constant except for archive size. These values are chosen based on desired feature discretization. We consider 4 different feature discretization values which include 5, 10, 15 and 20. Since we have three different features, this results in a total archive size of 125 (5*5*5), 1000, 3375 and 8000. From here on we denote all our MEHH variants as $MEHH_{<archive\_size>}$.

Both methods are evaluated using 31 independent runs. For each run we evaluate the final archive or population (in the case of GP) on the validation set and pick the best performing solution to be the representative for that run. This solution is then evaluated on the completely unseen test set. The final statistics given in Table \ref{table:gp_vs_map_elites_dev} are calculated based on the test performance of the 31 representative rules from each of the runs.

The results clearly indicate the benefit of using MEHH over GPHH in finding high performing priority rules for larger instances. For all archive sizes, our proposed MEHH presents better mean and median performance. $MEHH_{8000}$ performs the best with a 2.53\% improvement over GPHH. It is also more reliable in comparison to a GPHH due to low standard deviation in the results across multiple runs. In fact, all our MEHH variants have lower standard deviation in comparison to GPHH. We also see a correlation between increase in archive size and improvement in overall performance. This would make sense, since a larger archive would allow MEHH to maintain a larger set of useful genetic material that could greatly assist the evolutionary search. 

\begin{figure}[htbp!]
    \centering
    \includegraphics[width=9cm]{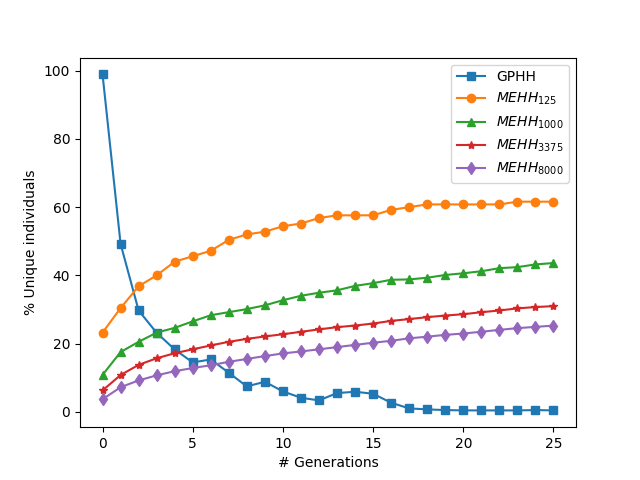}
    \caption{Change in diversity over generations for median runs of GPHH and MEHH.}
    \label{fig:diversity_plot}
\end{figure}

GPHH loses diversity over generations. Even if the solutions in GPHH are different in terms of the tree structure, they tend to be variants of one another resulting in the same performance. Essentially the bloat effect takes over resulting in longer trees with little to no improvement in terms of objective value. This is further illustrated in Figure \ref{fig:diversity_plot} which shows the percentage of unique solutions in the archives or the population for the respective median runs (based on test performance). As the run progresses, GPHH continues to lose diversity which greatly hinders its ability to find better solutions. MEHH on the hand continues to improve in terms of diversity as more and more cells in the archive get occupied. Table \ref{table:unique_individuals} sheds more light on this issue. It tabulates the number of unique individuals across all runs. In our experiments GPHH was run with a population size of 1024 yet on average it only has about 37 unique individuals at the end of a run. Even $MEHH_{125}$ manages to outperform GPHH in this aspect with more than twice as many unique individuals in the end. This again highlights the advantage and need for diversity preserving hyper-heuristics as early convergence, overfitting and bloat effect are all real issues affecting the quality of solutions in a standard GPHH.

%\textcolor{OliveGreen}{We also compare with the best performing arithmetic rule ($AR_{PSGS1}$) given in \cite{CHAND2018146} and demonstrate the performance improvement by MEHH.}
\begin{center}
\begin{table*}[htbp!]
\centering

\caption{Comparison  of test performance (average percentage deviation ) for GPHH and MEHH over all 31 runs. }
\label{table:gp_vs_map_elites_dev}
\begin{tabular}{ c c c c c c} 
 \hline
Measure  & GPHH & $MEHH_{125}$ & $MEHH_{1000}$ & $MEHH_{3375}$ & $MEHH_{8000}$\\ 
 \hline
Mean & 1005.93  & 1004.00 & 1003.61  & 1003.41  & 1003.40\\
Median & 1004.92  & 1003.42  & 1003.39  & 1003.22  & 1003.41\\
Best & 1002.93  & 1002.36  & 1001.72  & 1001.62  & 1001.47\\
Worst & 1010.15  & 1009.25  & 1006.34  & 1007.34  & 1004.64\\
St. Dev. & 2.01  & 1.50  & 0.87  & 1.25  & 0.80\\
 \hline 
\end{tabular}
\end{table*}
\end{center}

\begin{center}
\begin{table*}[h!]
\centering

\caption{Unique individuals at termination of training.}
\label{table:unique_individuals}
\begin{tabular}{ cccccc } 
 \hline
  & GPHH & $MEHH_{125}$ & $MEHH_{1000}$ & $MEHH_{3375}$ & $MEHH_{8000}$ \\ 
 \hline
Mean & 37 & 83 & 469 & 1124  & 1905\\
Median & 23 & 83 & 471  & 1124 & 1891\\ 
Min & 2 & 77 & 436  & 1045 & 1817\\
Max & 133 & 91 & 509  & 1190 & 2069\\
St. Dev. & 36 & 3 & 17  & 43 & 59\\
 \hline
\end{tabular}
\end{table*}
\end{center}
\subsection{MEHH Archive Coverage}

An important aspect to consider is the coverage of the MEHH grid. Coverage is calculated as the percentage of cells in the grid which contain a solution. Since it isn't necessary that a solution exists for each feature value, some cells will inevitably be empty in the grid.

\begin{center}
\begin{table}[H]
\centering
\caption{Coverage at termination for median run of MEHH for different grid sizes.}
\label{table:coverage}
\begin{tabular}{ c c }
 \hline
   Grid size & Coverage (\%) \\
  \hline
 $MEHH_{125}$ & 66.4 \\
$MEHH_{1000}$ & 47.1 \\
$MEHH_{3375}$ & 33.3 \\
$MEHH_{8000}$ & 23.6 \\
 \hline
\end{tabular}
\end{table}
\end{center}
A greater coverage is beneficial since it provides the decision maker with a greater number of choices. In our case, it allows us to have a better chance of avoiding over-fitted solutions. Table \ref{table:coverage} gives a summary of archive coverage by size for a typical run. We observe that archive coverage actually decreases with an increase in archive size. This may be due to a number of reasons. 

\begin{figure*}%
    \centering
    \subfloat[\centering $Feature_1$ vs $Feature_2$]{{\includegraphics[width=5cm]{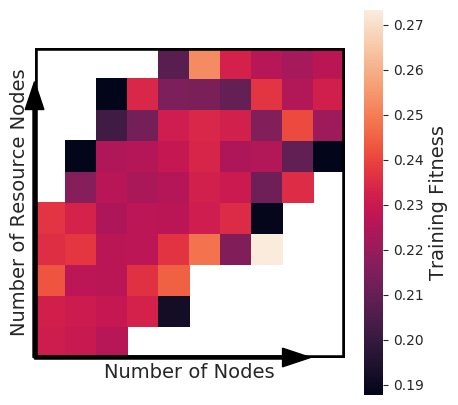} }}%
    \qquad
    \subfloat[\centering $Feature_2$ vs $Feature_3$]{{\includegraphics[width=5cm]{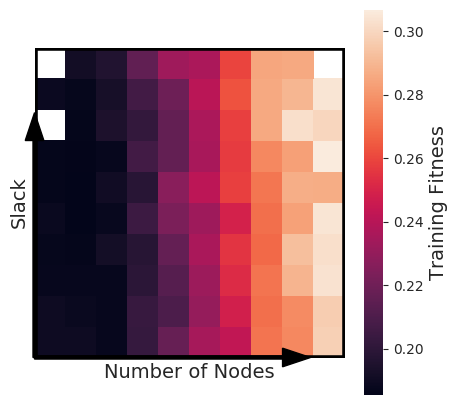} }}%
    \qquad
    \subfloat[\centering $Feature_2$ vs $Feature_3$]{{\includegraphics[width=5cm]{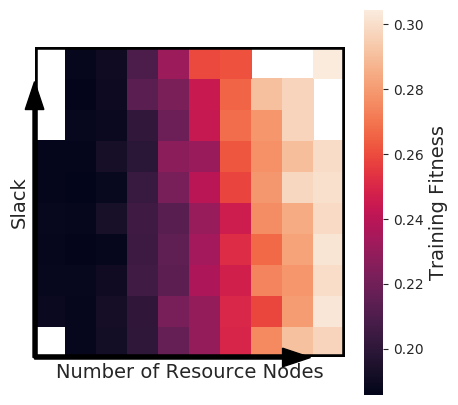} }}%
    \caption{Feature wise plots of the performance grid. The fitness of individuals corresponding to the third feature (in each of the plots) is averaged along the z axis}%
    \label{fig:performance_grid}%
\end{figure*}

First contributing factor is the correlation between two of the selected features. Number of nodes and number of resource nodes are positively correlated (Figure \ref{fig:performance_grid}). In fact one is a subset of the other and hence this means that certain feature combinations are realistically not possible. Furthermore, the same two features can only have integer values. This can create gaps in the feature space due to the non-continuous nature and once again lead to combinations or feature ranges which realistically cannot have any corresponding solutions. This problem can get worse with an increase in archive size. However, as noted earlier, we see a positive correlation between increase in archive size and improvement in results. Hence, even though the coverage for larger archive sizes is less, the quality is sufficient to ensure good results.

Figure \ref{fig:performance_grid} shows pair-wise feature-fitness plots. We can see that smaller trees generally have better training performance. This is a trend that has been reported in our previous work \cite{CHAND2019897} as well. Next we see that solutions with lower slack values also tend to perform better. This is again expected as less slack corresponds to more compact schedules with lower makespans.

\subsection{Comparison with existing rules}
Now that we have established that our MEHH outperforms a classical GPHH, we turn our attention to the quality of individual rules. It is important to analyze the performance of the evolved rules against existing human designed priority rules. Table \ref{table:priority_rules_performance} lists the performance of existing priority rules from literature on the test set. The best performing rule is MTS closely followed by LFT, FIFO and EST.

\begin{table}[htbp!]
\centering
\caption{Performance of human designed priority rules on test set.}
\label{table:priority_rules_performance}
\small
\begin{tabular}{  c c c  } 
 \hline
 Rule & Percentage Deviation & Makespan \\ 
 \hline
 EST & 1020.19 & 313714 \\
EFT & 1028.43 & 316192 \\
LST & 1020.43 & 313813 \\
LFT & 1014.39 & 311633 \\
SPT & 1077.04 & 330034 \\
FIFO & 1016.30 & 312323 \\
\textbf{MTS} & \textbf{1013.72} & \textbf{311336} \\
RAND & 1071.38 & 328539 \\
GRPW & 1044.91 & 323228 \\
GRD & 1081.61 & 331584 \\
IRSM & 1079.58 & 330329 \\
WCS & 1023.52 & 314897 \\
ACS & 1050.69 & 322928 \\

 \hline
\end{tabular}
\end{table}

As mentioned earlier, for each run we pick the best performing rule on the validation set to be the representative for that run and to be subsequently evaluated on the test set. From these 31 rules we select a sub-set of rules for further analysis and comparison with MTS. We pick 3 rules from GPHH and each of the MEHH variants. For each algorithmic variant we select the best, worst and median rules from the set of 31 representative rules selected across runs. These are of course selected based on test performance. Each of the MEHH rules are denoted as $MEHH_{<grid\ size >-<metric>}$ where \textit{metric} can be either B (Best), W(worst) or M(median) while the \textit{grid\_size} can be either 125, 1000, 3375, 8000. For GPHH, the rules are denoted as $GPHH_{<metric>}$. The performance of the selected rules is given in Table \ref{table:rules_testset}.

Based on the results, we can clearly see that all of our selected rules outperform the existing the human designed priority rules. Even in the worst case, our evolved rules are still better than MTS. Rule $MEHH_{8000-B}$ performs best in terms of average percentage deviation while $MEHH_{1000-B}$ performs the best in terms of cumulative makespan. In all cases, the rules selected from MEHH outperform the GPHH evolved rules. This indicates the strength of our proposed MEHH in discovering priority rules which perform exceptionally well on new unseen cases. 
\begin{center}
    \begin{table}[htp]
    \centering
    \small
    \caption{Performance of selected evolved rules on the test set.}
    \label{table:rules_testset}
    \begin{tabular}{ ccc } 
     \hline
     Rule & Deviation & Makespan \\
     \hline
   
%$AR_{PSGS1}$ \cite{CHAND2018146} & 1002.45 & 308472 \\
$GPHH_{B}$ & 1002.93 & 308460 \\ 
$MEHH_{125-B}$ & 1002.36 & 308510 \\ 
$MEHH_{1000-B}$ & 1001.72 & \textbf{308141} \\ 
$MEHH_{3375-B}$ & 1001.62 & 308199 \\ 
$MEHH_{8000-B}$ & \textbf{1001.47} & 308199 \\ 
$GPHH_{M}$ & 1004.92 & 309125 \\ 
$MEHH_{125-M}$ & 1003.42 & 309072 \\ 
$MEHH_{1000-M}$ & 1003.39 & 309000 \\ 
$MEHH_{3375-M}$ & 1003.22 & 308922 \\ 
$MEHH_{8000-M}$ & 1003.41 & 308958 \\ 
$GPHH_{W}$ & 1010.15 & 310597 \\ 
$MEHH_{125-W}$ & 1009.25 & 310290 \\ 
$MEHH_{1000-W}$ & 1006.34 & 309714 \\ 
$MEHH_{3375-W}$ & 1007.34 & 310190 \\ 
$MEHH_{8000-W}$ & 1004.64 & 309636 \\

         \hline
\end{tabular}
\end{table}
\end{center}
We continue our analysis by comparing the performance of the same set of rules on only hard instances from the test set. We characterise hard instances as those having RC $\geq$ 0.6 and RU $\geq$ 3. This essentially refers to instances where the resource requirements are high and the availability is relatively limited. The results for this comparison are given in Tables \ref{table:rules_hard_instances} and \ref{table:priority_rules_hard_instances}. We can see that once again MTS outperforms all other human designed priority rules. Similarly, $MEHH_{8000-B}$ performs best in terms of average percentage deviation while $MEHH_{3375-B}$ performs the best in terms of cumulative makespan. However, the gap in performance between the evolved rules and MTS is now much larger. This further emphasizes the usefulness of our evolved rules which perform equally well on new unseen difficult instances. For a potential user or decision maker, this information is handy as it gives an indication of how well the rules perform under difficult circumstances.

\begin{center}
    \begin{table}[htp]
    \centering
    \small
    \caption{Performance of selected evolved rules on hard instances. Hard instances were classified as having RC $\geq$ 0.6 and RU $\geq$ 3. }
    \label{table:rules_hard_instances}
    \begin{tabular}{ ccc } 
     \hline
     Rule & Deviation & Makespan \\
     \hline
   
%$AR_{PSGS1}$ & 2220.75 & 161653\\
$GPHH_{B}$ & 2219.69 & 161659 \\ 
$MEHH_{125-B}$ & 2218.42 & 161551 \\ 
$MEHH_{1000-B}$ & 2218.64 & 161576 \\ 
$MEHH_{3375-B}$ & 2217.96 & \textbf{161502} \\ 
$MEHH_{8000-B}$ & \textbf{2217.70} & 161523 \\ 
$GPHH_{M}$ & 2224.90 & 161998 \\ 
$MEHH_{125-M}$ & 2219.55 & 161684 \\ 
$MEHH_{1000-M}$ & 2220.31 & 161656 \\ 
$MEHH_{3375-M}$ & 2220.16 & 161656 \\ 
$MEHH_{8000-M}$ & 2220.33 & 161722 \\ 
$GPHH_{W}$ & 2234.72 & 162631 \\ 
$MEHH_{125-W}$ & 2230.40 & 162246 \\ 
$MEHH_{1000-W}$ & 2227.37 & 162196 \\ 
$MEHH_{3375-W}$ & 2227.69 & 162220 \\ 
$MEHH_{8000-W}$ & 2224.58 & 161935 \\ 
         \hline
\end{tabular}
\end{table}
\end{center}

Finally, we compare the performance of the same set of rules on the remaining PSPLib instances. As mentioned earlier, we used the entire J30 instance set for training. Here we compute the performance of these rules on the J60, J90 and J120 set. J60 and J90 both consist of 480 instances which are constructed using the same parameter combinations as J30. J120 consists of 600 instances which are constructed using a slighter harder set of parameter combinations \cite{KOLISCH1997205}. The results for this comparison are given in Tables \ref{table:rules_different_sets} and \ref{table:priority_rules_different_sets}. $MEHH_{8000-M}$ is the best performing rule for J60, while $MEHH_{1000-B}$ and $MEHH_{3375-M}$ perform the best on J90 and J120, respectively.  Majority of our evolved rules out-perform all of the human designed rules. The difference in performance is a lot smaller in this case but still significant if we consider the number of instances involved.

\begin{center}
    \begin{table}[h]
    \centering
    \small
    \caption{Performance of human designed priority rules on hard instances.}
    \label{table:priority_rules_hard_instances}
    \begin{tabular}{ ccc } 
     \hline
     Rule & Deviation & Makespan \\
     \hline

EST & 2242.00 & 163232 \\
EFT & 2251.00 & 163826 \\
LST & 2249.78 & 163724 \\
LFT & 2236.70 & 162772 \\
SPT & 2297.57 & 166841 \\
FIFO & 2237.51 & 162870 \\
\textbf{MTS} & \textbf{2234.50} & \textbf{162596} \\
RAND & 2284.17 & 166059 \\
GRPW & 2265.79 & 165272 \\
GRD & 2301.08 & 167284 \\
IRSM & 2297.44 & 166848 \\
ACS & 2255.45 & 164369  \\
WCS & 2250.37 & 163797  \\
         \hline
\end{tabular}
\end{table}
\end{center}
\begin{center}
    \begin{table}[h]
    \centering
    \small
    \caption{Performance of selected evolved rules on smaller instance sets. }
    \label{table:rules_different_sets}
    \begin{tabular}{ cccc } 
     \hline
     Rule & J60 & J90 & J120\\
     \hline
%$AR_{PSGS1}$ & \textbf{16.78} & 15.32 & \textbf{42.63} \\
$GPHH_B$ & 17.06 &  15.54 &   43.04 \\
$MEHH_{125-B}$ & 16.89 & 15.53 & 42.93  \\ 
$MEHH_{1000-B}$ & 16.97 & \textbf{15.26} & 43.21  \\ 
$MEHH_{3375-B}$ & 17.30 & 15.56 & 43.00  \\ 
$MEHH_{8000-B}$ & 17.09 & 15.62 & 43.16  \\ 
$GPHH_M$ & 16.94 & 15.46 & 42.99 \\
$MEHH_{125-M}$ & 17.08 & 15.49 & 43.08  \\ 
$MEHH_{1000-M}$ & 17.02 & 15.49 & 43.42  \\ 
$MEHH_{3375-M}$ & 16.87 & 15.42 & \textbf{42.83}  \\ 
$MEHH_{8000-M}$ & \textbf{16.82} & 15.52 & 42.88  \\ 
$GPHH_W$ & 17.32 & 15.63 & 43.33 \\
$MEHH_{125-W}$ & 17.74 & 15.54 & 43.31  \\ 
$MEHH_{1000-W}$ & 17.01 & 15.46 & 43.80  \\ 
$MEHH_{3375-W}$ & 16.88 & 15.50 & 43.39  \\ 
$MEHH_{8000-W}$ & 17.32 & 15.62 & 42.93  \\
\hline
\end{tabular}
\end{table}
\end{center}
\begin{center}
    \begin{table}[htbp!]
    \centering
    \small
    \caption{Performance of human designed priority rules on smaller instance sets.}
    \label{table:priority_rules_different_sets}
    \begin{tabular}{ cccc } 
     \hline
     Rule & J60 & J90 & J120\\
     \hline
EST & 21.68   & 21.45   & 55.78   \\
EFT & 22.46   & 21.82   & 55.77   \\
LST & \textbf{17.12}   & \textbf{15.80}   & 44.04   \\
LFT & 17.46   & 15.90   & \textbf{43.86}   \\
SPT & 23.77   & 23.60   & 60.33   \\
FIFO & 20.38   & 19.47   & 51.57   \\
MTS & 17.98   & 16.70   & 46.03   \\
RAND & 23.53   & 22.81   & 59.08   \\
GRPW & 22.32   & 21.83   & 57.87   \\
GRD & 24.11   & 23.18   & 61.30   \\
IRSM & 20.73   & 19.14   & 53.62   \\
ACS & 20.41   & 19.09   & 51.98   \\
WCS & 18.48   & 16.92   & 46.43   \\
\hline
\end{tabular}
\end{table}
\end{center}

\vspace{-2.0cm}
\subsection{Discussion}
The results presented above highlight the strength of our MEHH. MEHH produces rules which on average generalize significantly better to an unknown test set. GPHH starts with a diverse population which quickly erodes away as the evolutionary selection pressure starts filling up the population with copies of few 'good' solutions and their variants. MEHH on the other hand maintains an archive that has been partitioned based on selected features of interest. A new solution is only added into the archive if it exhibits features which the archive has not yet encountered or if it outperforms an existing solution with the same characteristics. This approach ensures and encourages a diverse population which in turn provides a wider range of options when picking the final representative rule. 

\begin{figure}[htbp!]
    \centering
    \includegraphics[width=9cm]{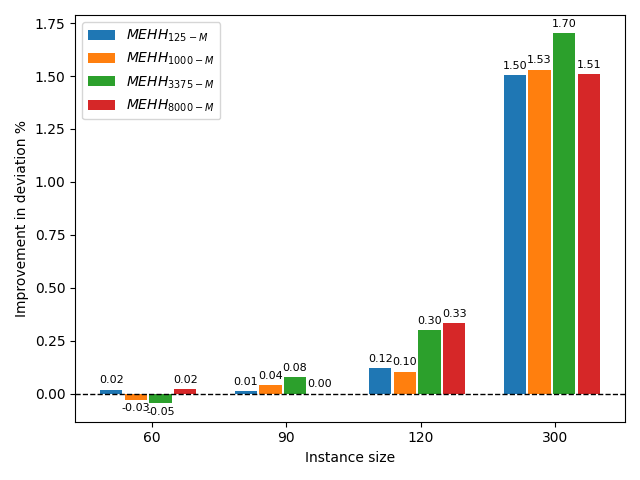}
    \caption{Improvement in percentage deviation given by MEHH rules over $GPHH_M$.}
    \label{fig:gp_vs_map_elites}
\end{figure}

We observed the same pattern when comparing individually selected rules. Rules selected from MEHH always outperformed rules selected from GPHH in the best, median and worst case scenarios. An illustration of this is given in Figure \ref{fig:gp_vs_map_elites}. The gap in performance increases with an increase in instance size. As we tackle larger more difficult instances, the benefits of MEHH evolved rules become more apparent. Our evolved rules also outperform existing human designed rules by a significant margin. Prior to this, majority of the studies on hyper-heuristic focused solely on smaller instances (30 to 120 activities). For the first time, we presented an in-depth study with larger and more difficult instances. 

\begin{figure}[htbp!]
    \centering
    \includegraphics[width=10cm]{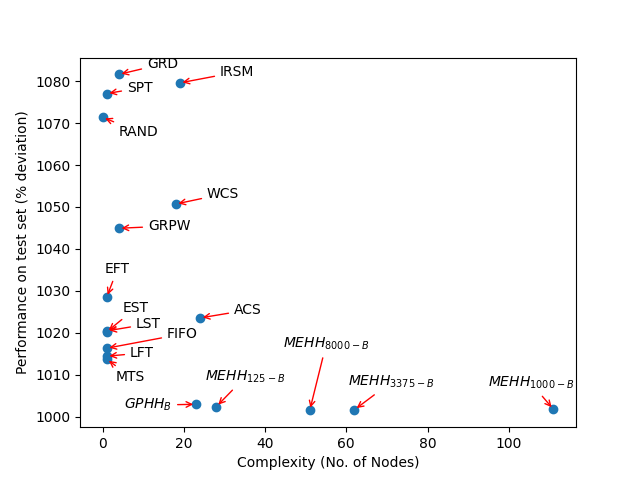}
    \caption{Comparison of evolved and existing rules.}
    \label{fig:complexity_plot}
\end{figure}

The rules evolved in this study will be highly desirable in dynamical scheduling environments where sudden and unexpected changes require quick scheduling responses. In such situations the goal is to get a 'good enough' solution in a reasonable amount of time as opposed to focusing on optimality. Furthermore the evolved rules can also be used to seed or augment existing meta-heuristic \cite{FANG2012890} and exact methods to improve their performance in cases where optimality and not speed is the desired goal. Some of the rules discovered in this study are simplified and presented in Table \ref{table:generated_rules}. A further comparison in terms of performance and rule complexity is given in Figure \ref{fig:complexity_plot}. Rule $MEHH_{125-B}$ offers the best trade-off between performance and complexity. In fact, except for $MEHH_{1000-B}$, all other evolved rules exhibit a reasonable level of complexity and can be easily utilized by decision makers. 

Furthermore, MEHH allows the decision maker to customize the search process to their liking by choosing features that are more relevant to their scheduling environment. For instance, in some real world applications incorporating slack is an important aspect of the scheduling process. This is done in order to 'cushion' the schedule from unplanned disruptions. The decision maker can thus refer to the archive and pick a heuristic that appropriately reflects their desired quality indicators(eg. a decent trade-off between performance and slack).

\begin{center}
    \begin{table*}[htbp!]
    \centering
    \small
    \caption{Selected evolved rules and their performance on the test set. }
    \label{table:generated_rules}
    \begin{tabular}{ ccc } 
     \hline
     Rule & Description & Deviation \\
     \hline
     \\
    $GPHH_{B}$ & \pbox{10cm}{-AvgRReq - EF - ES - LF - 2*LS - Max(AvgRReq, TSC) - Min(AvgRReq, -TSC) + Min(EF, LS) } & 1002.93\\
    \\
    $MEHH_{125-B}$ & \pbox{10cm}{-2*AvgRReq + EF - LF - Max(EF, TSC) + Max(LS, RR/TPC - Max(MaxRReq + MinRReq, MaxRReq + TSC)) } & 1002.36 \\
    \\
    $MEHH_{1000-B}$ & \pbox{10cm}{-AvgRReq + LF*LS - TSC + (EF*MaxRReq - MaxRReq*TPC)*(LS*MaxRReq + MaxRReq/TSC)*(-MaxRReq*RR - 1/(LS*MaxRReq))*(-MaxRReq*Min(AvgRReq, EF) + Min(ES, RR) + Min(Min(1, AvgRReq*TSC)/(EF*TPC + 2*MinRReq), 1/(-AvgRReq - Min(AvgRReq, TPC))) + Min(1/(MinRReq + RR), Max(TPC, TSC)))*Max(ES*LF, Min(LS, RR))*Min(ES*MinRReq, Max(EF, MinRReq))*Min(MinRReq + TSC, Max(MinRReq, RR))*Min(AvgRReq, TSC, Max(EF, LS))} & 1001.72 \\
    \\
    $MEHH_{3375-B}$ & +\pbox{10cm}{Max(LS, MinRReq) + Min(-TSC*(AvgRReq*TSC + AvgRReq + ES + 1 + 1/(-EF - LS)), ES**2*LS*MinRReq*Max(AvgRReq, EF, AvgRReq*ES) + Min(-MinRReq, MaxRReq*RR*(AvgRReq + 1/MaxRReq), LS - RR) - 1/(EF*ES*(EF + MaxRReq)))} & 1001.62 \\
    \\
    $MEHH_{8000-B}$ & \pbox{10cm}{LF*LS - Max(-LF, Min(AvgRReq, MaxRReq)) - Min(-AvgRReq, AvgRReq)*Min(LF, -MaxRReq)*Min(-LS, ES + MaxRReq) + Min(-TSC, 2*LF*(-RR - TPC + Min(MinRReq, RR))*Min(AvgRReq, EF))} & 1001.47 \\

     \hline
\end{tabular}
\end{table*}
\end{center}
\vspace{-1cm}
\section{Conclusion}
In this study we explored the use of MAP-Elites based Hyper Heuristics (MEHH) for evolving priority rules for the classical RCPSP. A systematic investigation was presented comparing MEHH with a standard GPHH. MEHH outperformed GPHH by maintaining a diverse repertoire of solutions which allowed for better generalization on new unseen instances. In particular, this study focused on larger instances which have often been ignored/under-studied in the existing literature. The rules discovered through MEHH were extensively compared with human designed priority rules. The comparisons revealed major performance gaps and even cases where the worst case outputs from MEHH performed better than the best human designed rules. 

In future research we will consider improving the proposed MEHH by investigating the impact of each of the individual features and also considering newer more advanced features. We will also consider applying our proposed MEHH to dynamic problem instances and evaluating its efficacy. 

\bibliography{references.bib}

\end{document}